# Exploring Cumulative Effects in Survival Data Using Deep Learning Networks

Kang-Chung Yang and Shinsheng Yuan

Institute of Statistical Science, Academia Sinica, Taipei, Taiwan

Email: syuan@stat.sinica.edu.tw

## Abstract

In epidemiological research, modeling the cumulative effects of time-dependent exposures on survival outcomes presents a challenge due to their intricate temporal dynamics. Conventional spline-based statistical methods, though effective, require repeated data transformation for each spline parameter tuning, with survival analysis computations relying on the entire dataset, posing difficulties for large datasets. Meanwhile, existing neural network-based survival analysis methods focus on accuracy but often overlook the interpretability of cumulative exposure patterns. To bridge this gap, we introduce CENNSurv, a novel deep learning approach that captures dynamic risk relationships from time-dependent data. Evaluated on two diverse real-world datasets, CENNSurv revealed a multi-year lagged association between chronic environmental exposure and a critical survival outcome, as well as a critical short-term behavioral shift prior to subscription lapse. This demonstrates CENNSurv's ability to model complex temporal patterns with improved scalability. CENNSurv provides researchers studying cumulative effects a practical tool with interpretable insights.

## Introduction

In occupational epidemiology and pharmacoepidemiology, understanding the influence of prolonged exposure to risk factors on survival outcomes has long been a central focus of research. These studies often address the complexity of time-varying exposures and their cumulative effects on the risk of adverse events. For instance, studies have examined how cumulative radiation exposure over decades impacts lung cancer mortality among uranium miners [1] and how prolonged medication use leads to fall-related hospitalizations among the elderly [2]. These cumulative effects, influenced by temporal patterns, challenge researchers to fully capture their dynamic nature. This pursuit inspires researchers to explore diverse approaches in survival analysis to better understand the interplay between exposure histories and survival outcomes.

Over the past few decades, statistical methods have evolved to enhance our understanding of these dynamic relationships. Early models relied on fixed aggregation schemes to summarize past exposures, which often oversimplified their temporal complexity. Later, more flexible techniques were introduced, such as time-weighted functions [3, 4] and spline-based methods integrated into the Cox proportional hazards framework. One notable example is the Weighted Cumulative Exposure (WCE) method [5], which has been widely applied in pharmacoepidemiological studies [6]. Spline-based approaches have also led to the development of distributed lag non-linear models (DLNM), which provide a unified framework to represent both exposure-response and lag-response relationships simultaneously through a bidimensional cross-basis structure [7]. DLNMs have also been utilized in cohort studies, including the uranium miners study [1], to assess the temporal impact of exposure histories on survival outcomes. Achieving successful outcomes with DLNMs requires the precise selection of basis functions, which critically shapes the interpretation of exposure and lag effects. This process is further complicated by interactions with lag length settings, rendering the tuning of predefined basis functions highly challenging. For large datasets, the repeated data transformation for each spline parameter adjustment, followed by survival analysis across the entire dataset, creates a substantial computational burden. This motivates exploring deep learning networks, which can learn complex cumulative relationships directly from time-dependent data, bypass the constraints of predefined basis functions in spline-based methods.

In recent years, deep learning has gained traction in survival analysis, primarily for the purpose of improving survival outcome predictions. Early models, such as DeepSurv [8], DeepHit [9], and Cox-Time [10] from the PyCox framework, focus on time-fixed survival data. DeepSurv extends the Cox model with neural networks while retaining standard Cox proportional hazard loss, DeepHit models survival distributions for competing risks using alternative loss functions to improve accuracy, and Cox-Time extends Cox regression beyond the proportional hazards to support time-varying effects. Later models, such as Dynamic-DeepHit [11] and DNNSURV [12], handle time-dependent survival data. Dynamic-DeepHit incorporates longitudinal data and competing risks, and DNNSURV transforms survival analysis into a regression problem using pseudo values. However, these approaches, whether for time-fixed or time-dependent data, are designed for predictive performance. They lack the neural architectures necessary to capture cumulative exposure effects over time. This limits their ability to explore the time-varying relationships that are central to our research.

To address this gap, we introduce CENNSurv (Cumulative Effect modeling via Neural Networks in SURVival analysis), a novel deep learning approach that enhances the Cox model framework with neural networks. CENNSurv learns complex cumulative relationships directly from time-dependent data, avoiding the constraints of predefined basis functions in spline-based methods. This provides a scalable, data-driven solution for modeling cumulative effects. Unlike prior methods, CENNSurv employs specialized layers to model the impact of evolving exposure through weighted past data. This enables CENNSurv to assess time-varying influences, offering a distinct advantage in analyzing shifting risk patterns.

We demonstrate CENNSurv's practical utility by applying it to two real-world datasets: one involving long-term occupational exposure (e.g., radon exposure in uranium miners) and another involving dynamic behavioral patterns (e.g., listening behaviors in subscription churn). These examples validate CENNSurv's ability to handle time-dependent data and provide interpretable insights.

# Method

## Modeling cumulative effects in time-dependent survival data

In survival analysis, cumulative effects arise from time-dependent covariates that vary dynamically rather than remaining static. These covariates, such as exposure levels, change over time and require dynamic modeling to reflect their evolving impact on survival risk. To account for this, we index covariates by time $t$. Time-based indexing enables exposure history $X = \{x_t\}$, $t = 1, 2, \ldots, T$, to vary, where the time index $t$ represents discrete points that correspond to real-world time intervals (e.g., absolute calendar times or relative times from the start of the observation). Typically, the time span is evenly divided into segments, each of which is assigned a unique index $t$.

To quantify past exposures' impact on event risk, we define $h(X, t)$ as the cumulative exposure effect at time $t$, integrating the entire exposure history:

$$h(X, t) = \sum_{l=0}^{L} f(x_{t-l}) w(l)$$

This formulation represents $h(X, t)$ as a convolution function, where past exposure effects, $f(x_{t-l})$ are weighted by $w(l)$ over a range of delays. Here, $l = 0, 1, \ldots, L$ is the delay time index, indicating discrete steps into the past, and $L$ is the window size in the convolution. It is determined as an empirical parameter based on the study's temporal dynamics (e.g., the optimal lag period for capturing exposure effects). Here, $t$ and $l$ are integers, allowing $t - l$ to denote prior time points in the exposure history. In this formula, $f(x_{t-l})$ denotes the exposure effect at a prior time, $t - l$, and $w(l)$ captures the influence of varying delays on current risk.

Traditionally, spline basis functions model $f(.)$ and $w(.)$ to create smooth exposure- and lag-response representations. In these representations, $f(.)$ is the exposure function that yields $f(x_{t-l})$, and $w(.)$ is the lag function that yields $w(l)$. However, our deep learning approach directly estimates $h(X, t)$, avoiding predefined transformations and dynamically adapting to complex, nonlinear patterns during training. This approach is beneficial when $f(.)$ and $w(.)$ exhibit discontinuities or sharp transitions because it allows the model to adapt to scenarios in which cumulative effects are absent, such as when only current effects are present. For interpretability, we assume independent decomposition of $f(.)$ and $w(.)$, enabling separate analysis of exposure intensity and time delays.

The hazard function is then modeled using Cox's proportional hazards model:

$$\lambda(t|X) = \lambda_0(t) \exp[h(X, t)]$$

Here, $\lambda(t|X)$ denotes the hazard function, $\lambda_0(t)$ the baseline hazard, and $h(X, t)$ the log-hazard from cumulative exposure.

## Overview of the Deep Learning Approach

The model is designed to directly learn the exposure function $f(.)$ and the lag function $w(.)$ to estimate the cumulative exposure effect $h(X, t)$, thus avoiding the limitations of traditional spline-based methods. First, we model the exposure effect $f(x)$ using a residual dense block. Then we use a convolutional layer to model the lag effect $w(l)$. This approach adapts to intricate patterns in the data and incorporates structural constraints to address identifiability issues, ensuring robust, interpretable results. The resulting log-hazard $h(X, t)$ is derived from these learned functions. The following sections describe the input data structure, model architecture, and training procedure in detail.

**Inputs:** Each subject's exposure history is represented as an input matrix of dimensions $(N, T, C)$, where $N$ is the number of subjects, $T$ represents discrete time points, and $C$ is the number of covariates. For the purposes of this analysis, we will assume $C = 1$.

**Outputs and Loss Function:** The network outputs the log-hazard function, $h(X, t)$, which represents the risk of an event at time $t$, given the covariates $X$. Event and risk masks, which are derived from the observed event times and the at-risk subjects, restrict the survival likelihood computation to the relevant subjects. We optimize $h(X, t)$ using the negative log partial likelihood loss from DeepSurv, extended with Efron's method to handle tied events:

$$l(\theta) := -\frac{1}{m} \sum_{t:d_t \neq 0} \left[ \sum_{i \in D(t)} \hat{h}_\theta(X_i, t) - \sum_{j=1}^{d_t} \log \left( \sum_{i \in R(t)} e^{\hat{h}_\theta(X_i, t)} - \frac{j-1}{d_t} \sum_{i \in D(t)} e^{\hat{h}_\theta(X_i, t)} \right) \right]$$

In this context, $\hat{h}_\theta(X_i, t)$ denotes the log-hazard estimation for subject $i$ at time $t$; $D(t)$ represents the event set at time $t$ and comprises subjects who experience the event at that specific time; $R(t)$ is the risk set at time $t$; $d_t$ is the number of events occurring at time $t$; and $m$ is the number of events in one batch of the full dataset.

**Identifiability Problems in Cumulative Effect Modeling**

In our approach, we model the log-hazard $\hat{h}_\theta(X, t)$ using a neural network to estimate $h(X, t)$, representing the cumulative exposure-response effects over time. Although this formulation enables flexible learning of exposure-response patterns, further decomposition is necessary to separate $f(.)$ and $w(.)$, as these two components represent different aspects of the cumulative effect.

In the process of deconvolving $h(.)$ into its components $f(.)$ and $w(.)$, however, two key identifiability problems arise:

**Multiplicative Associativity in the Convolution Operation:** The convolution operation between $f(x_{t-l})$ and $w(l)$ exhibits multiplicative associativity. This means that the resulting function, $h(X, t)$, can shift or scale in ways that make uniquely separating $f(x_{t-l})$ and $w(l)$ diffcult. This issue can be expressed as follows:

$$h(X, t) = \sum_{l=0}^{L} [kf(x_{t-l})][k^{-1}w(l)]$$

Here, $k$ is a scaling factor, and its inverse $k^{-1}$ ensures that the product $h(X, t)$ remains unchanged. This leads to non-unique estimates of the exposure effect, $f(x_{t-l})$, and the lag effect, $w(l)$.

**Log-Hazard Shifting in the Partial Log-Likelihood Calculation:** During the calculation of the partial log-likelihood, estimated log-hazard values within the same risk set can be shifted by a constant. This means that the hazard for all individuals at risk can be scaled simultaneously, making it difficult to interpret the results in absolute terms.

$$l(\theta) := -\frac{1}{m} \sum_{t:d_t \neq 0} \left[ \sum_{i \in D(t)} \widehat{h}_\theta(X_i, t) - \sum_{j=1}^{d_t} \log \left( \sum_{i \in R(t)} e^{\widehat{h}_\theta(X_i,t)} - \frac{j-1}{d_t} \sum_{i \in D(t)} e^{\widehat{h}_\theta(X_i,t)} \right) \right]$$

$$= -\frac{1}{m} \sum_{t:d_t \neq 0} \left[ \sum_{i \in D(t)} \widehat{h}_\theta(X_i, t) - \widehat{h}_\theta(0, t) - \sum_{j=1}^{d_t} \log \left( \sum_{i \in R(t)} e^{\widehat{h}_\theta(X_i,t) - \widehat{h}_\theta(0,t)} - \frac{j-1}{d_t} \sum_{i \in D(t)} e^{\widehat{h}_\theta(X_i,t) - \widehat{h}_\theta(0,t)} \right) \right]$$

To address these issues, we designed the network architecture to enforce constraints that ensure identifiability during training.

### Detailed Network Architecture

Figure 1 illustrates the network's overall architecture, highlighting the sequence of residual dense blocks and the integration of the Conv1D layer with time-dependent data.

### Exposure Function Modeling with Residual Dense Blocks

We model the exposure function, $f(.)$, using a residual connection defined as $f(x) = h(x) + x$, where $h(x)$ is a residual component learned by the model. Our goal is to ensure the identifiability constraint $f(0) = 0$, which prevents log-hazard shifting issues. To achieve this constraint, we define $h(x) = g(x) - g(0)$, where $g(x)$ is a trainable function learned via a residual dense block, and $g(0)$ is its output when the input is a zero tensor (i.e., no exposure). This output is computed with frozen parameters to eliminate variability from batch normalization and dropout. Combining these definitions, the exposure function becomes $f(x) = g(x) - g(0) + x$, which allows us to train $g(x)$ to obtain the desired $f(x)$.

### Lag Function Modeling with Convolutional Layer

The output, $f(x)$, from the residual dense block is processed by a one-dimensional convolutional layer (Conv1D) to model the lag function, $w(.)$. This layer uses a single filter to produce a scalar output per time point. The kernel size is equal to the lag length, which captures cumulative effects over a time window. Causal padding ensures that the output at time $t$ depends only on the current and past time points. The layer has no bias term and uses linear activation to preserve the raw convolution output. It also applies a unit L2 norm constraint ($\|w\|_2 = 1$)) to the kernel to normalize the weights, prevent scale ambiguity, and stabilize the optimization process.

### Training Procedure

The dataset is divided into training and testing sets at a ratio of 90/10. Stratified sampling based on survival times and events to preserves consistency in the Kaplan-Meier (K-M) distribution across the sets. We use cross-validation on the training set to tune hyperparameters, selecting those that accommodate model complexity and ensure generalization. Then, we train a series of models with varying levels of smoothness regularization on the Conv1D layer and assess their performance. We adjust this regularization in a subjective, exploratory manner to observe how $w(l)$ evolves under different smoothness constraints. We avoid treating smoothness as a hyperparameter because doing so would lead to selecting the model with the lowest loss, typically one without smoothness constraints, rather than allowing us to explore all smoothness levels. We choose the final model based on metrics from the test set, including loss and C-index, as well as the interpretability of the deconvoluted $f(x)$ and $w(l)$. This ensures a balance of effectiveness and explainability.

## Simulation Study

### Generation of Survival Data

We simulated data for 5,000 individuals, each with 100 days of exposure history. We generated exposure patterns using predefined functions: exposure functions $f(.)$, which took linear and nonlinear (plateau) forms to model exposure effects; and lag functions $w(.)$, which included continuous (decay) and discontinuous (current, stepwise) patterns to capture lagged effects. The exposure values $x$ were normalized to the range $[0, 1]$, and the lag $l$ ranged from 0 to 20 days. For each individual on each day, we computed the log-hazard $h(X, t)$, where $h(X, t) = \sum_{l=0}^{L} f(x_{t-l})w(l)$, as defined in the Method section. This resulted in a sequence of log-hazard values over 100 days. We also defined the exposure-lag contribution, $f(x)w(l)$, which represents the effect of exposure $x$ at lag $l$. By using evenly spaced $x$ and $l$ values as inputs, we can create a three-dimensional exposure-lag contribution surface that simultaneously visualizes the shapes of $f(.)$ and $w(.)$. Figure 2 shows the exposure function $f(.)$, the lag function $w(.)$, and the 3D exposure-lag contribution surface, with the z-axis representing $f(x)w(l)$. To ensure that the survival data reflected time-dependent risks, we applied the permAlgo algorithm [13], which uses a permutation-based approach to assign user-specified survival times and events. This process maintains a consistent link between the log-hazard values and the survival outcomes.

### Analysis and Performance Evaluation with CENNSurv

After generating exposure histories and the corresponding survival information, we analyzed the data using CENNSurv to model the complex relationships between exposure, lag, and survival outcomes. This analysis followed the procedures outlined in the Methods section. To explore the impact of regularization, we trained CENNSurv models with varying smoothness strengths and evaluated their performance. In parallel, we applied the DLNM framework to the same data, testing linear and B-spline functions (of degrees 2 and 3) for $f(.)$ and $w(.)$, which yielded nine combinations. We then selected the final model based on the smallest BIC using standard Cox proportional hazard models. To assess the accuracy of the deconvoluted $f(.)$ and $w(.)$, we defined a metric called GMSE (Grid Mean Squared Error), which quantifies the discrepancy between the true and predicted exposure-lag contributions by creating a grid of these contributions and calculating the mean squared error across it. We evaluated performance using 200 randomly generated datasets. We compared CENNSurv models of different smoothness strengths against the final DLNM model via two-sided independent t-tests. The performance metrics, averaged across these datasets for scenarios S1–S4, are summarized in Table 1 and Supplementary Figures S1 – S4.

### Results of the Simulation Study

Table 1 shows the simulation results for CENNSurv in four scenarios (S1–S4). Performance metrics (loss, C-index, and GMSE) are averaged across 200 simulated datasets and reported as means with standard deviations in parentheses. These results are then compared against those of DLNM.

In Scenario S1 (linear-current), CENNSurv outperformed DLNM in terms of loss, C-index, and GMSE under no smoothing, strength 1, and strength 5, showing significant improvement. However, as the smoothing strength increased, the performance metrics declined slightly, showing no significant difference from DLNM at strength 10. In Scenario S2 (Plateau-Current), CENNSurv outperformed DLNM at no smoothing and low smoothing strengths (1 and 5), exhibiting lower loss, a higher C-index, and competitive GMSE values. However, as smoothing strength increased to 10, performance declined significantly, with the C-index dropping to 0.6497 (vs. 0.6573) and GMSE rising to 0.1478 (vs. 0.1357), indicating notable underperformance due to over-regularization. In Scenario S3 (plateau-decay), CENNSurv had comparable loss and C-index values to DLNM, but a higher GMSE, despite a reduction in GMSE with increased smoothing strength (e.g., decreasing from 0.0064 to 0.0027 at strength 10 versus DLNM's 0.0006). In Scenario S4 (plateau-stepwise), CENNSurv showed significant improvements in loss and C-index compared to DLNM, as well as notably lower GMSE (e.g., 0.0048 at strength 10 versus DLNM's 0.0158), indicating an advantage in deconvolution accuracy.

CENNSurv exhibits strong performance in scenarios S1, S2, and S4 at lower smoothing strengths, effectively addressing discontinuous patterns and sharp breakpoints that pose challenges to the spline-based approach of DLNM. However, our analysis revealed that excessive smoothing in S2 led to degraded performance, as indicated by rising GMSE values. In scenario S3, CENNSurv delivers comparable loss and C-index values to DLNM, showcasing its adaptability, while DLNM's notably low GMSE is likely due to its use of fewer parameters, limiting the flexibility of deconvoluted functions across diverse datasets. Despite GMSE reductions with stronger smoothing, CENNSurv's values remained higher than this benchmark.

Supplementary Figures S1-S4 illustrate the deconvoluted $f(.)$ and $w(.)$ functions, and the differences in exposure-lag contributions across smoothing strengths. These visualizations reveal how CENNSurv captures patterns in S1 and S2 at optimal smoothing, the impact of excessive smoothing in these scenarios, how CENNSurv reasonably fits S3's smooth patterns, and its deconvolution advantage in S4. Based on these findings, we applied CENNSurv to real-world datasets, including those from the Colorado Plateau uranium miners cohort and the KKBox churn prediction data, to explore its practical utility.

## Real-world applications

**Colorado Plateau Uranium Miners Cohort**

This well-established epidemiological study examines the long-term health impacts of radon exposure on lung cancer risk among miners in the region. The dataset is publicly available on GitHub [14]. It includes information on 3,347 miners, 258 of whom died of lung cancer. Key demographic variables include age at study entry and exit, age during mining, and cumulative radon exposure recorded every five years. Following the preprocessing steps from the DLNM paper, we treated annual radon exposure as a time-dependent variable to input into our model.

We trained CENNSurv across a range of smoothness strengths, from none to a maximum strength of 10. The results are depicted in Figure 3(a)–(d). In the model with no smoothness constraints, we observed a distinct inflection point at $x = 0.15$ in the exposure function, suggesting a potential threshold effect; however, the lag function lacked a clear, interpretable trend. Additionally, the bootstrap-derived 95% confidence intervals exhibited high variability and asymmetry, indicating possible extreme values in the dataset. As smoothness strength increased, however, a trend gradually emerged in the lag function, revealing a positive weighting between lag5 and lag25 at strength 5. This indicates a strong association between lung cancer mortality and radon exposure from 5 to 25 years prior.

A comparative analysis with DLNM is presented in Supplementary Figure S5, displaying the three-dimensional exposure-lag contribution surfaces of CENNSurv (smoothness strength = 5) and DLNM side by side. This comparison shows that, despite using a neural network to automatically fit nonlinear exposure and lag functions, CENNSurv achieves results comparable to DLNM, which relies on tuned spline basis functions. Nevertheless, the cumulative effect of radon exposure is not apparent in the raw data, necessitating adequate smoothness to reveal underlying risk patterns. Model performance metrics are detailed in Table 2.

**KKBox's Churn Prediction Challenge**

This study is based on a dataset from a 2017 Kaggle data analysis competition. KKBox, a leading Asian music streaming platform, provided the dataset to facilitate research on customer churn prediction. The dataset includes both static user attributes (e.g., gender, registration city, and registration date) and dynamic behavioral data (e.g., transaction history and daily listening records). Due to the extensive user base of the dataset and its two years of continuous observational data, survival analysis was conducted to determine if specific listening behavior patterns influence the risk of churn. According to the competition guidelines, churn occurs when "the user decides not to renew their subscription (no new valid service subscription within 30 days after the current membership expires)." We identified 10,213 customers predicted to churn and examined their listening histories, which spanned up to 790 days. Daily listening times were normalized to the range [0, 1], and a lag period of 40 days was selected based on preliminary experiments to balance capturing sufficient past engagement patterns and maintaining computational efficiency.

After the CENNSurv training process, our model successfully estimated the exposure and lag functions. Figure 4 shows the deconvoluted exposure and lag functions. The results indicate that subscribers who actively use the service have a lower churn risk (with an approximate log hazard of -1.5), while variations in daily listening duration do not significantly impact churn risk. Additionally, the deconvoluted lag function reveals a notable trend: a negative weight is observed on the 29th day before churn, followed by a sharp reversal to a positive weight on the 30th day. Multiplying the lag function by the exposure function yielded risk estimates for different listening durations across varying lag periods. This led to the 3-D exposure-lag contribution surface visualization in Figure 5. The 3-D exposure-lag contribution surface slice plots reveal a clear relationship between subscriber churn and listening duration. Specifically, listening behavior on the 29th day before churn is associated with an increased churn risk. In contrast, listening on the 30th day before churn has the opposite effect and acts as a negative risk factor. This reversal pattern suggests that users with minimal listening activity between days 40 and 30 before churn who resume listening between days 29 and 20 before churn face the highest churn risk on the decision day.

With the estimated risk dynamics from the exposure and lag functions, we now analyze user segmentation to investigate how churn manifests across different subscriber groups. To better understand churn dynamics, we performed a cluster analysis on daily listening histories. Among the 10,213 subscribers who churned, we selected 7,218 individuals with observation periods longer than 180 days. We applied Dynamic Time Warping (DTW) to quantify the similarity of their listening behaviors.

Subscribers were grouped into five clusters (Figure 6), with Clusters 3, 4, and 5 accounting for 96% of the population. Despite differences in average listening duration (approximately 2.5, 1.2, and 0.3 hours, respectively), these clusters exhibit consistent behavioral trends. Specifically, their listening activity declines rapidly from day 60 before churn and nearly ceases by day 30. However, a sudden resurgence in listening activity is observed on day 29, followed by a sharp decline leading to churn. This pattern is likely driven by KKBox's subscription mechanism. Subscribers with extended observation periods are typically stable users. Those who decide not to renew their subscription, however, gradually reduce their listening activity. When their subscription expires, users receive reminders and are granted temporary free access, which may encourage them to renew. This results in a temporary spike in listening activity. Nevertheless, users who do not convert back to paid subscribers are ultimately classified as churned.

Figures 7 and Table 3 summarize the risk dynamics across clusters. The model successfully captures the sharp rise in risk as churn approaches. Specifically, Clusters 3, 4, and 5 experienced relative increases in risk of 7.77%, 9.79%, and 62.69%, respectively, on the churn decision day compared to 60 days prior. Clusters 1 and 2 exhibit relatively stable listening behavior, reflected by their minimal or negative relative risk changes at churn. This suggests that these users were less influenced by the observed re-engagement pattern. Model performance metrics are detailed in Table 4.

## Discussions

We developed CENNSurv, a deep learning-based survival analysis method that effectively captures cumulative effects in time-dependent survival data. In simulation studies, CENNSurv demonstrated superior adaptability with discontinuous exposure-lag patterns (e.g., S1, S2, and S4) under appropriate smoothing constraints. It achieved better loss, C-index, and GMSE than traditional DLNM. In the continuous S3 scenario, CENNSurv delivered comparable loss and C-index values to DLNM, reflecting its versatility across diverse settings. Real-world applications, such as the Colorado Plateau uranium miners cohort and KKBox churn prediction, confirmed CENNSurv's ability to identify lagged associations (e.g., a five- to 25-year lag for lung cancer mortality) and critical behavioral transitions (e.g., days 29–30 before churn). These applications provide interpretable and potentially actionable insights.

CENNSurv serves as a gateway to survival analysis using neural networks for researchers studying cumulative effects. It models various time-lagged associations with fewer assumptions than the DLNM method, facilitating quick pattern identification and flexible lag length selection while reducing the need for complex manual optimization due to spline and lag length interactions. As long as the lag length encompasses the true signal, the resulting patterns remain consistent, minimizing tuning efforts. Furthermore, CENNSurv's batch training approach enhances scalability by processing data in portions.

Despite its strengths, CENNSurv has notable limitations compared to traditional statistical methods. First, parameter tuning is computationally intensive and often requires a significant amount of time to achieve optimal performance, which offsets some of its efficiency gains. Second, final model selection relies on manual intervention and lacks a convenient metric, such as BIC. Third, the assumption of independent exposure and lag functions to enhance interpretability may oversimplify complex interactions. Lastly, although CENNSurv supports multiple input variables, identifiability issues among variables may cause nonlinear variables to be simultaneously deconvolved, resulting in information concentrating in one variable and degrading the performance of the others.

Future research should focus on three key areas to address the limitations of CENNSurv and its current challenges. First, developing an automated metric, such as a BIC equivalent, will allow for the data-driven selection of models that balance performance and interpretability. This reduces the need for manual observation and decision-making across different smoothing constraints. Second, advancing efficient hyperparameter tuning methods will mitigate the high computational intensity and streamline the optimization process. Third, improving multi-variable modeling capabilities through additional constraints, hierarchical modeling, or regularization techniques will enhance CENNSurv's capacity to handle intricate interactions. These advancements will bridge the gap between statistical and neural network approaches, expanding CENNSurv's potential as a versatile tool for studying cumulative effects across domains.

# Figures and Tables

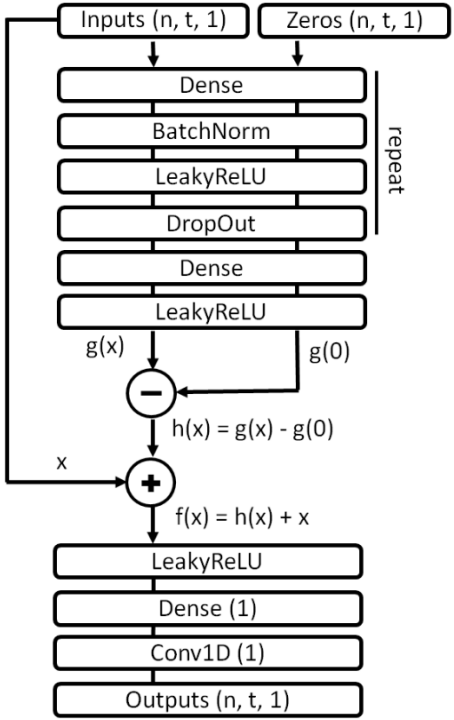

**Figure 1. Network Architecture Overview.** This figure shows the network architecture, emphasizing the residual dense block and the integration of the Conv1D layer with time-dependent data.

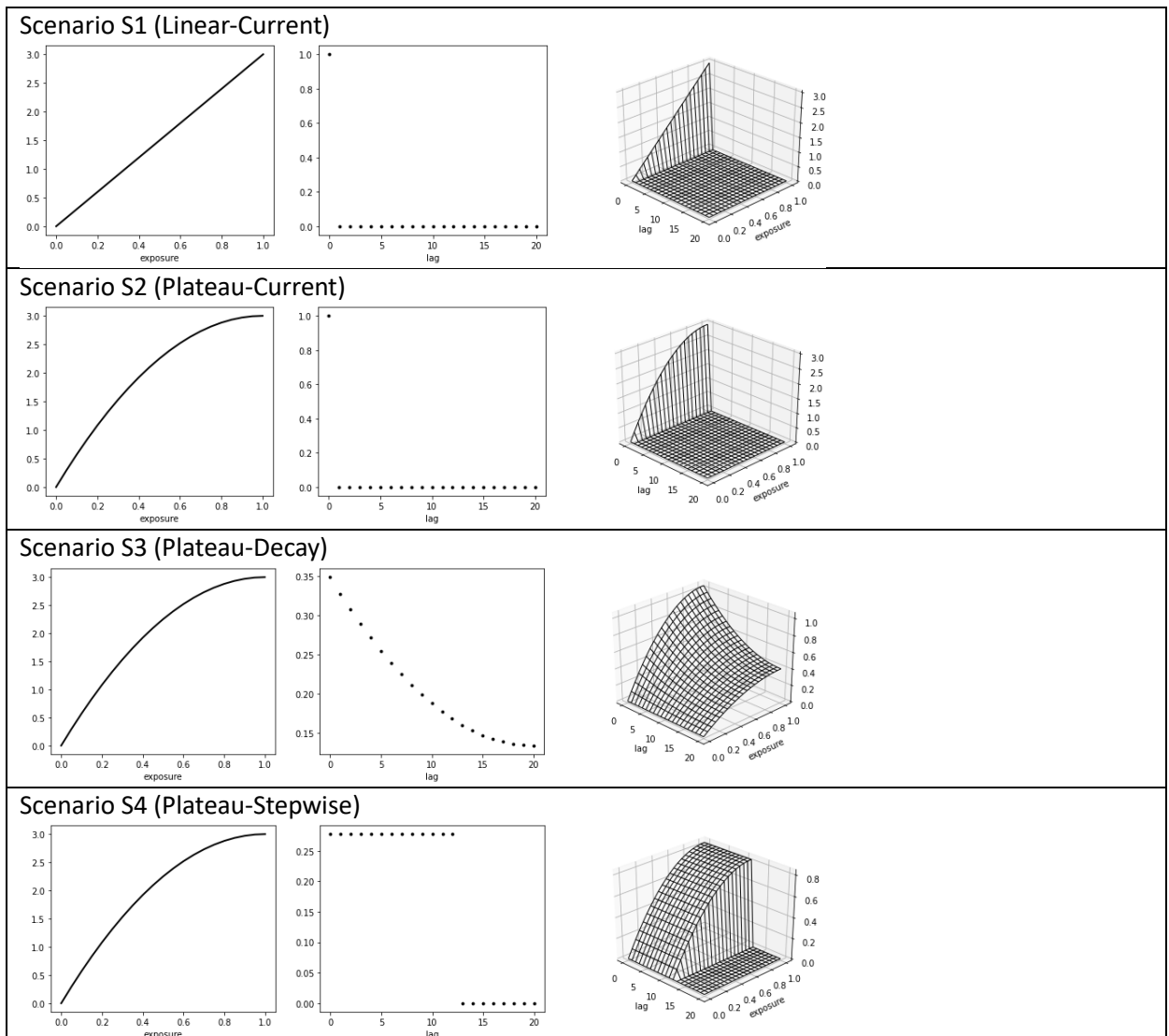

**Figure 2. True Exposure and Lag Functions Across Scenarios.** This figure shows the true exposure function $f(.)$, true lag function $w(.)$, and three-dimensional exposure-lag contribution surface for four simulation scenarios: S1 (Linear-Current), S2 (Plateau-Current), S3 (Plateau-Decay), and S4 (Plateau-Stepwise). Each scenario has distinct temporal dynamics, with $x$ sampled uniformly over [0, 1] for exposure and $l$ over [0, 20] for lag.

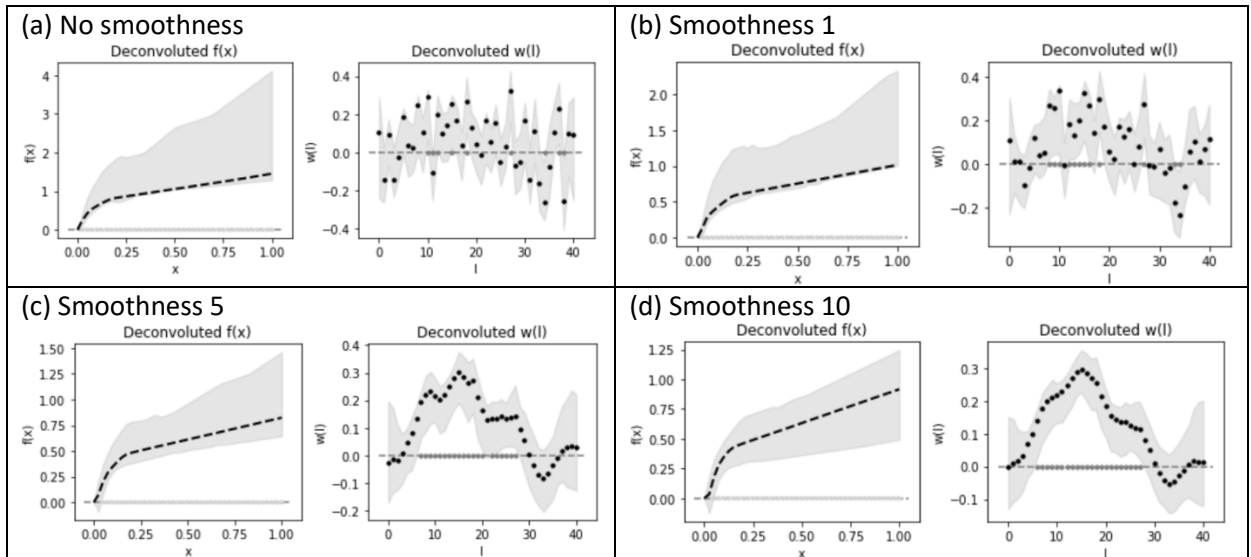

**Figure 3. Deconvoluted Exposure and Lag Functions.** This figure shows the deconvoluted exposure and lag functions with varying degrees of smoothness constraints. Panel (a): No smoothness constraint. The exposure function exhibits a clear inflection point at x = 0.15, while the lag function lacks a clear trend. Panels (b) through (d): Increasing smoothness constraints (strength = 1, 5, 10). The lag function stabilizes and reveals strong positive weighting between lag5 and lag25. The shaded gray areas represent the 95% confidence intervals obtained via bootstrapping with 100 iterations.

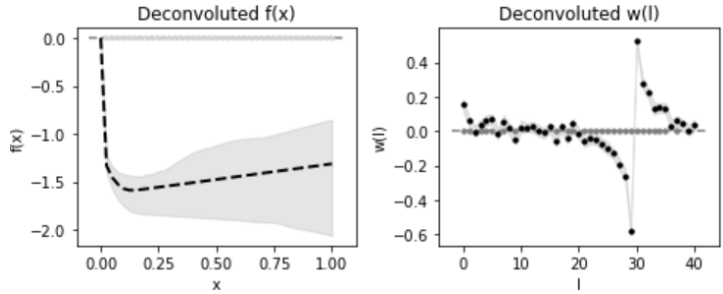

**Figure 4. Deconvoluted Exposure and Lag Functions.** This figure presents the deconvoluted exposure and lag functions. The left panel shows that subscribers who engage with the platform have a lower churn risk. Meanwhile, listening duration does not significantly impact risk variation. The right panel depicts the lag function and reveals a pronounced reversal pattern in weights between days 20 and 40 before churn. The shaded gray areas indicate the 95% confidence intervals obtained by bootstrapping 100 iterations of the original data.

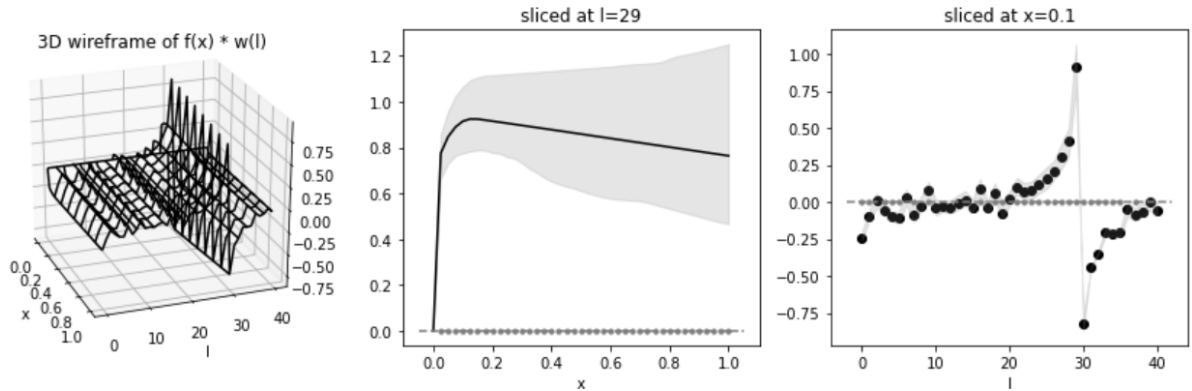

**Figure 5. Three-Dimensional Exposure-Lag Contribution Surface and Log Hazard Slice Plots.** This figure shows the estimated log hazard of churn across different lag periods. The left panel presents the log hazard response surface, and the right panel displays slice plots highlighting variation in churn risk on key days before churn.

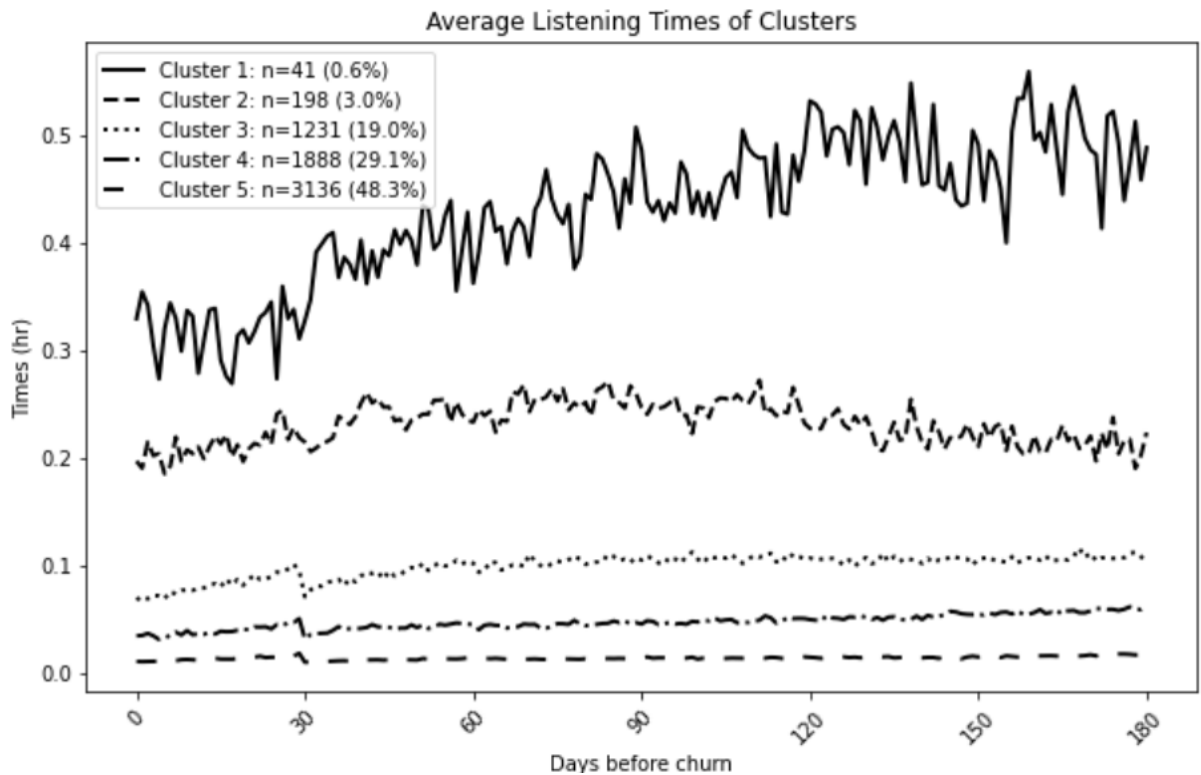

**Figure 6. Cluster-Based Listening Behavior Patterns.** This figure shows the average daily listening behavior of five subscriber clusters during the 180 days before churn. Each line represents a different cluster and illustrates variations in listening engagement over time. Clusters 3, 4, and 5 notably exhibit a distinct re-engagement pattern around day 29 before churn, while Clusters 1 and 2 demonstrate relatively stable listening behavior.

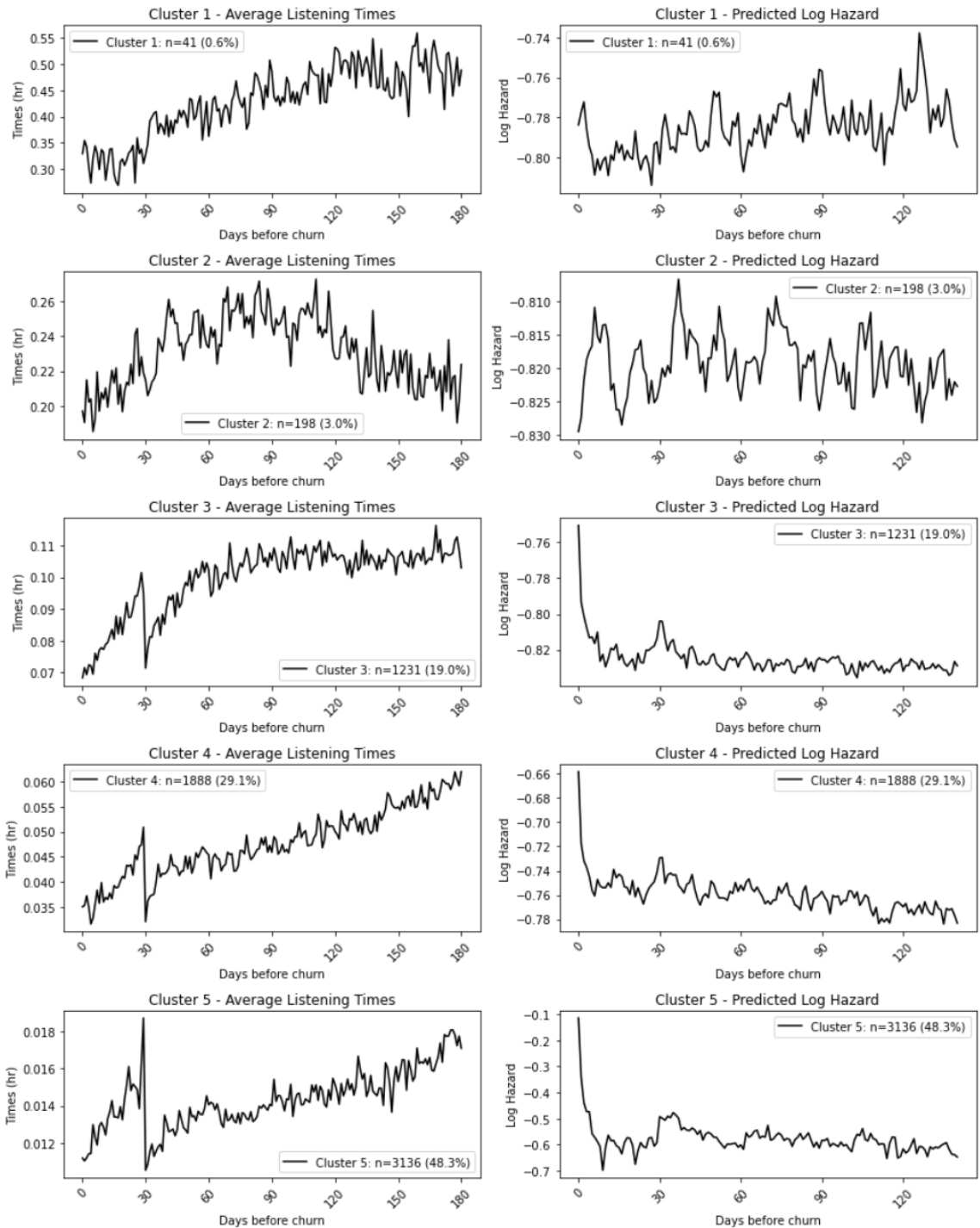

**Figure 7. Cluster-Specific Log Hazard Trends.** This figure contains two panels. The left panel shows the daily listening behavior of five clusters as individual time series, revealing variations in listening engagement. The right panel shows the corresponding log hazard trends and demonstrates the alignment between the re-engagement pattern and increasing churn risk for clusters 3, 4, and 5 during the final ten days before churn.

**Table 1. Performance Comparison of CENNSurv and DLNM Across Simulation Scenarios.** This table presents simulation results for CENNSurv and DLNM across scenarios S1–S4, evaluated over 200 datasets. Means are reported, with standard deviations in parentheses. A two-sided independent t-test assessed significance, comparing the results of CENNSurv at each smoothing strength (no smoothing, strengths 1, 5, and 10) against DLNM, with * indicating better performance ($p < 0.05$) and † indicating worse performance ($p < 0.05$) for CENNSurv relative to DLNM.

|  | CENNSurv No Smoothness | CENNSurv Smoothness 1 | CENNSurv Smoothness 5 | CENNSurv Smoothness 10 | DLNM |
|---|---|---|---|---|---|
| **Loss** | | | | | |
| S1 (Linear-Current) | 4.8286* (0.0381) | 4.8471* (0.0511) | 4.8951* (0.0628) | 4.9275 (0.0636) | 4.9352 (0.0312) |
| S2 (Plateau-Current) | 4.8987* (0.0322) | 4.9169* (0.0451) | 4.9688* (0.0477) | 4.9917 (0.0533) | 4.9846 (0.0276) |
| S3 (Plateau-Decay) | 4.5259 (0.0459) | 4.5247 (0.0458) | 4.5234 (0.0457) | 4.5242 (0.0453) | 4.5194 (0.0452) |
| S4 (Plateau-Stepwise) | 4.5176* (0.0501) | 4.5153* (0.0498) | 4.5154* (0.0495) | 4.5173* (0.0501) | 4.5359 (0.0473) |
| **C-index** | | | | | |
| S1 (Linear-Current) | 0.7199* (0.0142) | 0.7131* (0.0200) | 0.6944* (0.0269) | 0.6809 (0.0294) | 0.6822 (0.0156) |
| S2 (Plateau-Current) | 0.6889* (0.0144) | 0.6824* (0.0186) | 0.6617* (0.0258) | 0.6497† (0.0301) | 0.6573 (0.0159) |
| S3 (Plateau-Decay) | 0.7818 (0.0122) | 0.7820 (0.0122) | 0.7821 (0.0122) | 0.7821 (0.0123) | 0.7826 (0.0122) |
| S4 (Plateau-Stepwise) | 0.7812* (0.0127) | 0.7814* (0.0128) | 0.7814* (0.0127) | 0.7812* (0.0127) | 0.7780 (0.0126) |
| **GMSE** | | | | | |
| S1 (Linear-Current) | 0.0032* (0.0013) | 0.0194* (0.0321) | 0.0618* (0.0366) | 0.0817 (0.0241) | 0.0784 (0.0049) |
| S2 (Plateau-Current) | 0.0128* (0.0065) | 0.0440* (0.0554) | 0.1302 (0.0402) | 0.1478† (0.0392) | 0.1357 (0.0046) |
| S3 (Plateau-Decay) | 0.0064† (0.0026) | 0.0034† (0.0029) | 0.0027† (0.0021) | 0.0027† (0.0023) | 0.0006 (0.0007) |
| S4 (Plateau-Stepwise) | 0.0064* (0.0024) | 0.0034* (0.0027) | 0.0038* (0.0035) | 0.0048* (0.0050) | 0.0158 (0.0006) |

**Table 2. Model Performance Metrics for CENNSurv and DLNM on the Uminers Dataset.** This table shows the loss values and C-index for CENNSurv and DLNM under different configurations on the Colorado Plateau uranium miners cohort dataset. CENNSurv achieves its best performance at smoothness 5, while its performance at smoothness 10 is comparable to that of DLNM with tuned knot positions.

| Uminers | Loss | C-index |
|---|---|---|
| **CENNSurv** | | |
| No smoothness | 4.4452 | 0.7335 |
| Smoothness 1 | 4.4399 | 0.7385 |
| Smoothness 5 | 4.4176 | 0.7412 |
| Smoothness 10 | 4.4292 | 0.7370 |
| | | |
| **DLNM** | | |
| Default (B-spline degree=2) | 4.5615 | 0.6983 |
| Tuned knot positions | 4.4423 | 0.7365 |

**Table 3. Summary Statistics of Cluster-Specific Risk Patterns on the KKBox Dataset.** This table provides a quantitative summary of cluster-specific listening behaviors and risk dynamics. Clusters 3, 4, and 5 show substantial risk increases.

| | Cluster 1 | Cluster 2 | Cluster 3 | Cluster 4 | Cluster 5 |
|---|---|---|---|---|---|
| **N (%)** | 41 (0.63%) | 198 (3.05%) | 1231 (18.96%) | 1888 (29.07%) | 3136 (48.29%) |
| **Listening time on the churn decision day (hrs)** | 7.91 | 4.73 | 1.64 | 0.84 | 0.27 |
| **Listening time on the 60th day before churn (hrs)** | 8.70 | 5.60 | 2.46 | 1.09 | 0.34 |
| **Log hazard on the churn decision day** | -0.78 | -0.83 | -0.75 | -0.66 | -0.11 |
| **Log hazard on the 60th day before churn** | -0.80 | -0.82 | -0.83 | -0.75 | -0.60 |
| **Relative Risk Increase (%)** | 1.52 | -0.46 | 7.77 | 9.79 | 62.69 |

**Table 4. Model Performance Metrics for CENNSurv and DLNM on the KKBox Dataset.** This table shows the loss values and C-index for CENNSurv and DLNM on the KKBox dataset.

| KKBox | Loss | C-index |
|---|---|---|
| **CENNSurv** | | |
| No smoothness | 5.6641 | 0.6821 |
| | | |
| **DLNM** | | |
| Default (B-spline degree=2) | 8.0904 | 0.6404 |

# Supplementary Material

Kang-Chung Yang and Shinsheng Yuan

Institute of Statistical Science, Academia Sinica, Taipei, Taiwan

Email: syuan@stat.sinica.edu.tw

Scenario S1 (Linear-Current)

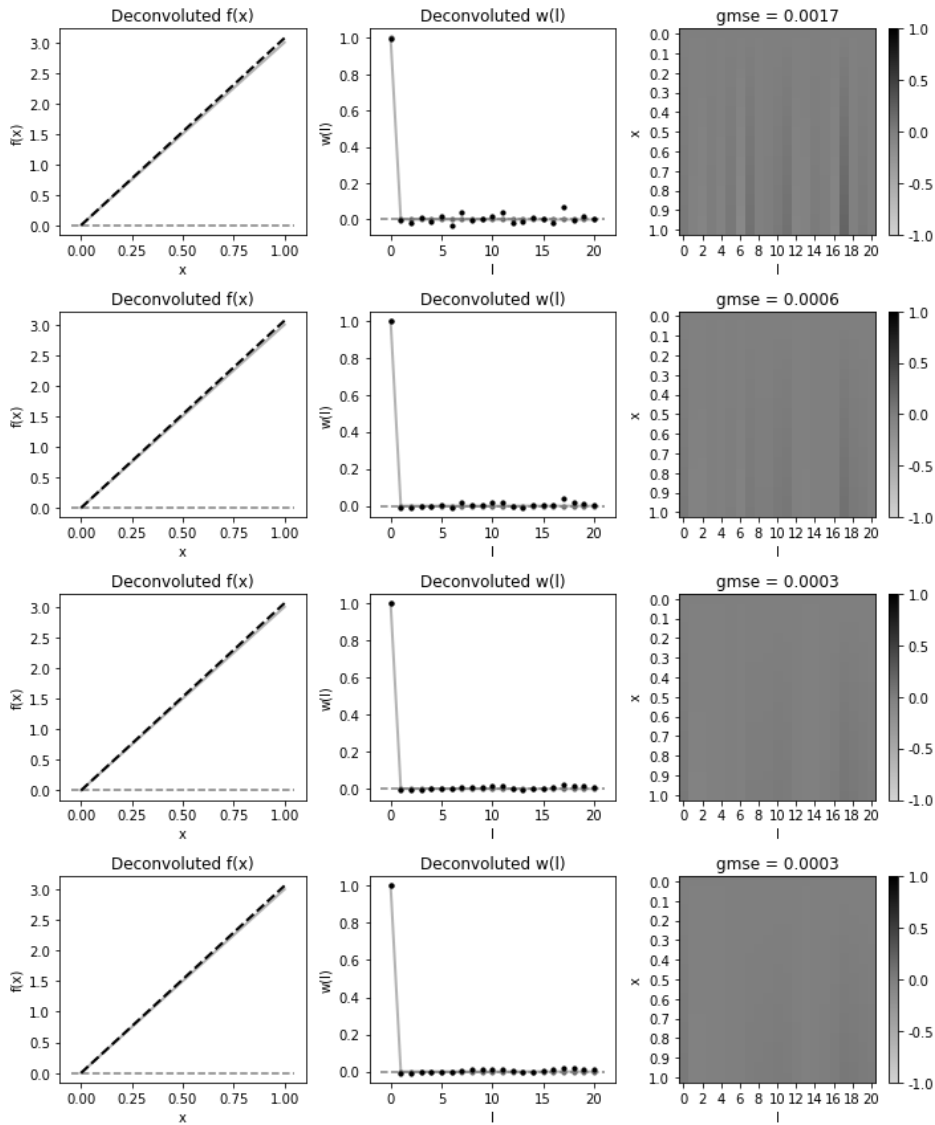

**Supplementary Figure S1.** This figure illustrates results for Scenario S1 (Linear-Current) using CENNSurv, with each row representing a different smoothing parameter (no smoothing, and strengths of 1, 5, and 10). Column 1 shows the deconvoluted exposure function, Column 2 displays the deconvoluted lag function, with gray solid lines representing true functions and black dashed lines with dots indicating

deconvoluted predicted functions from a single simulation. Column 3 compares true and predicted exposure-lag contributions in grid form. Note that this is an illustrative figure, showing results from one of 200 simulations.

Scenario S2 (Plateau-Current)

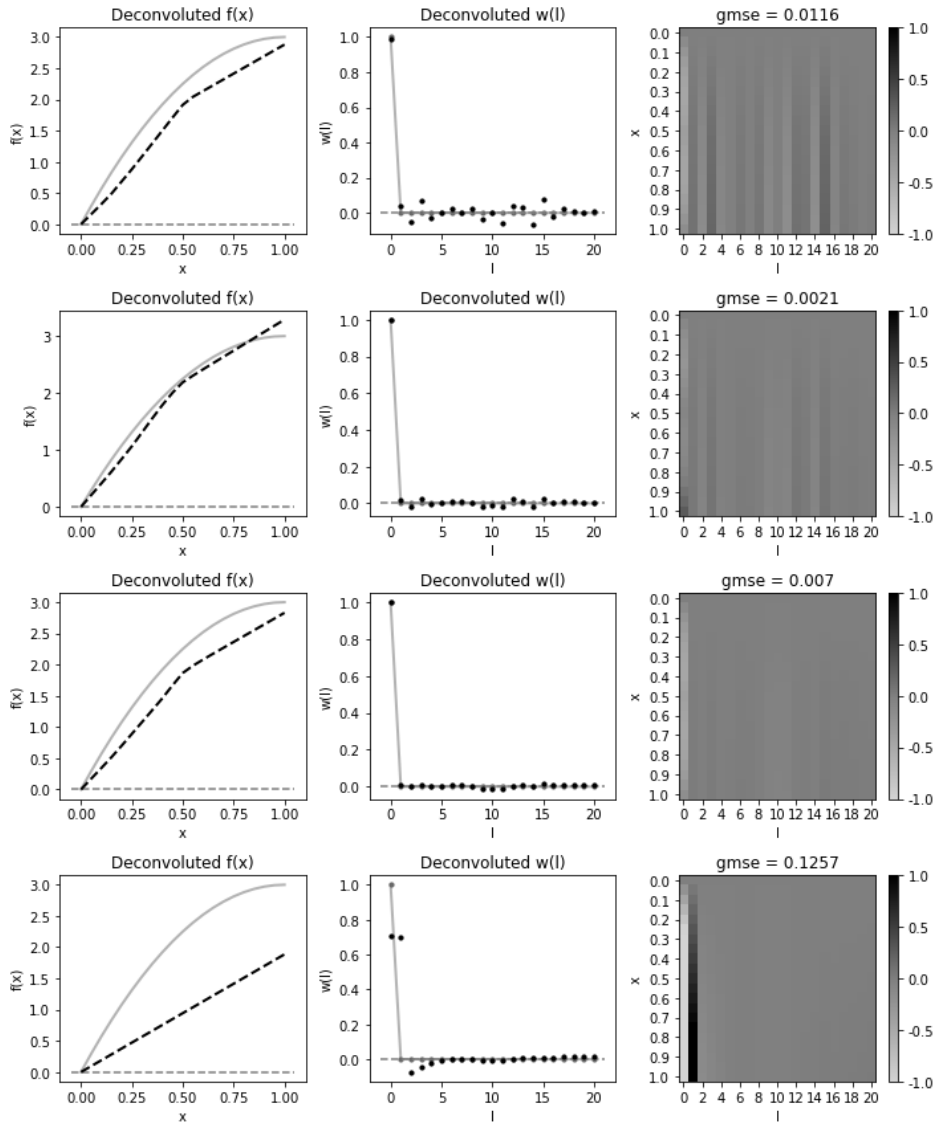

**Supplementary Figure S2.** This figure illustrates results for Scenario S2 (Plateau-Current) using CENNSurv, with each row representing a different smoothing parameter (no smoothing, and strengths of 1, 5, and 10). Column 1 shows the deconvoluted exposure function, Column 2 displays the deconvoluted lag function, with gray solid lines representing true functions and black dashed lines with dots indicating deconvoluted predicted functions from a single simulation. Column 3 compares true and predicted exposure-lag contributions in grid form. Note that this is an illustrative figure, showing results from one of 200 simulations.

Scenario S3 (Plateau-Decay)

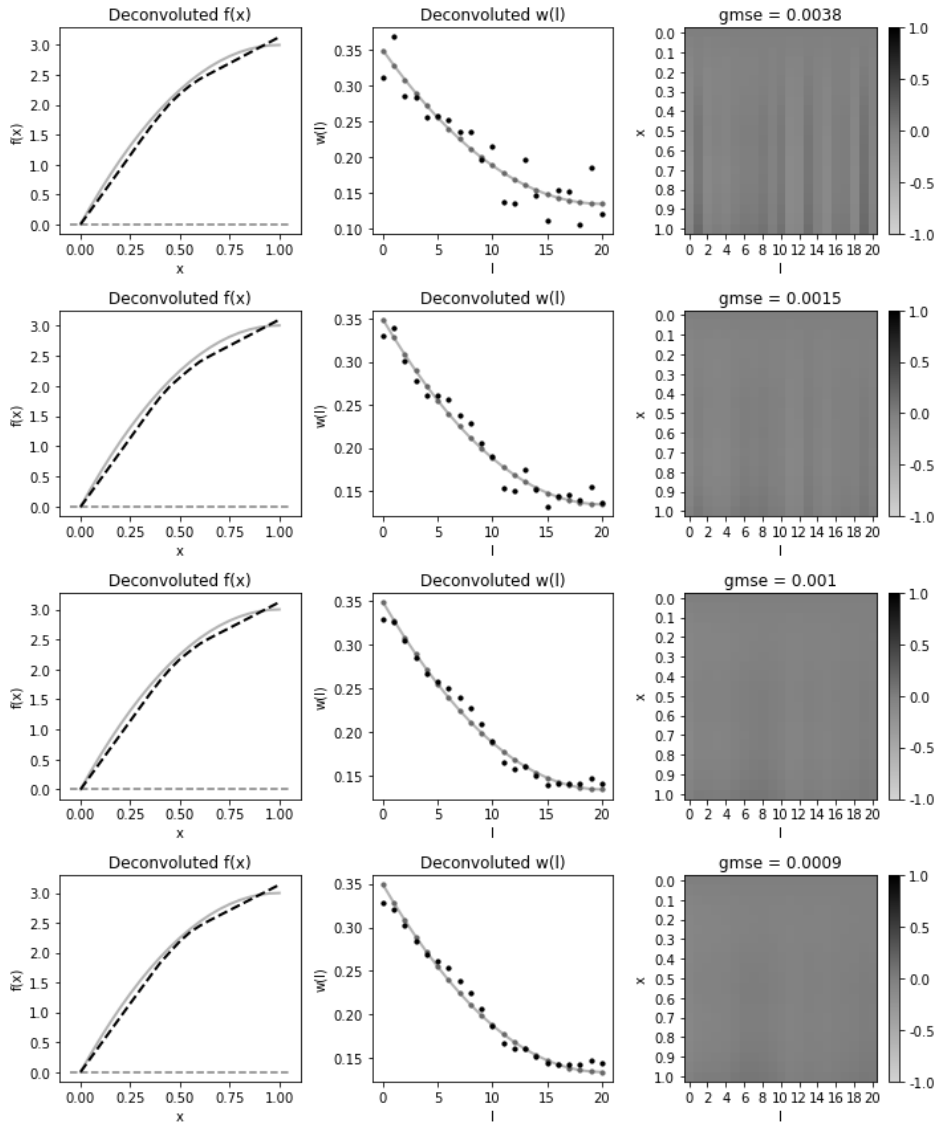

**Supplementary Figure S3.** This figure illustrates results for Scenario S3 (Plateau-Decay) using CENNSurv, with each row representing a different smoothing parameter (no smoothing, and strengths of 1, 5, and 10). Column 1 shows the deconvoluted exposure function, Column 2 displays the deconvoluted lag function, with gray solid lines representing true functions and black dashed lines with dots indicating deconvoluted predicted functions from a single simulation. Column 3 compares true and predicted exposure-lag contributions in grid form. Note that this is an illustrative figure, showing results from one of 200 simulations.

Scenario S4 (Plateau-Stepwise)

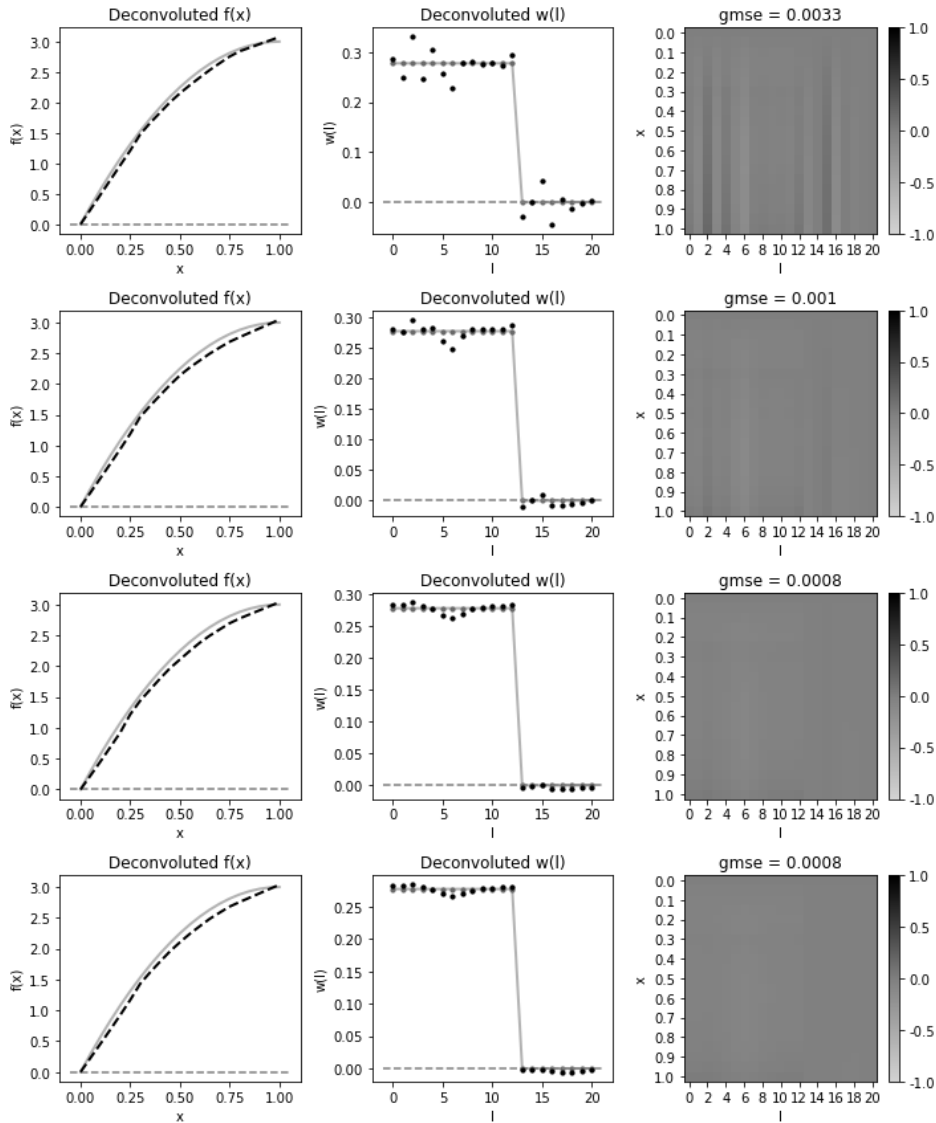

**Supplementary Figure S4.** This figure illustrates results for Scenario S4 (Plateau-Stepwise) using CENNSurv, with each row representing a different smoothing parameter (no smoothing, and strengths of 1, 5, and 10). Column 1 shows the deconvoluted exposure function, Column 2 displays the deconvoluted lag function, with gray solid lines representing true functions and black dashed lines with dots indicating deconvoluted predicted functions from a single simulation. Column 3 compares true and predicted exposure-lag contributions in grid form. Note that this is an illustrative figure, showing results from one of 200 simulations.

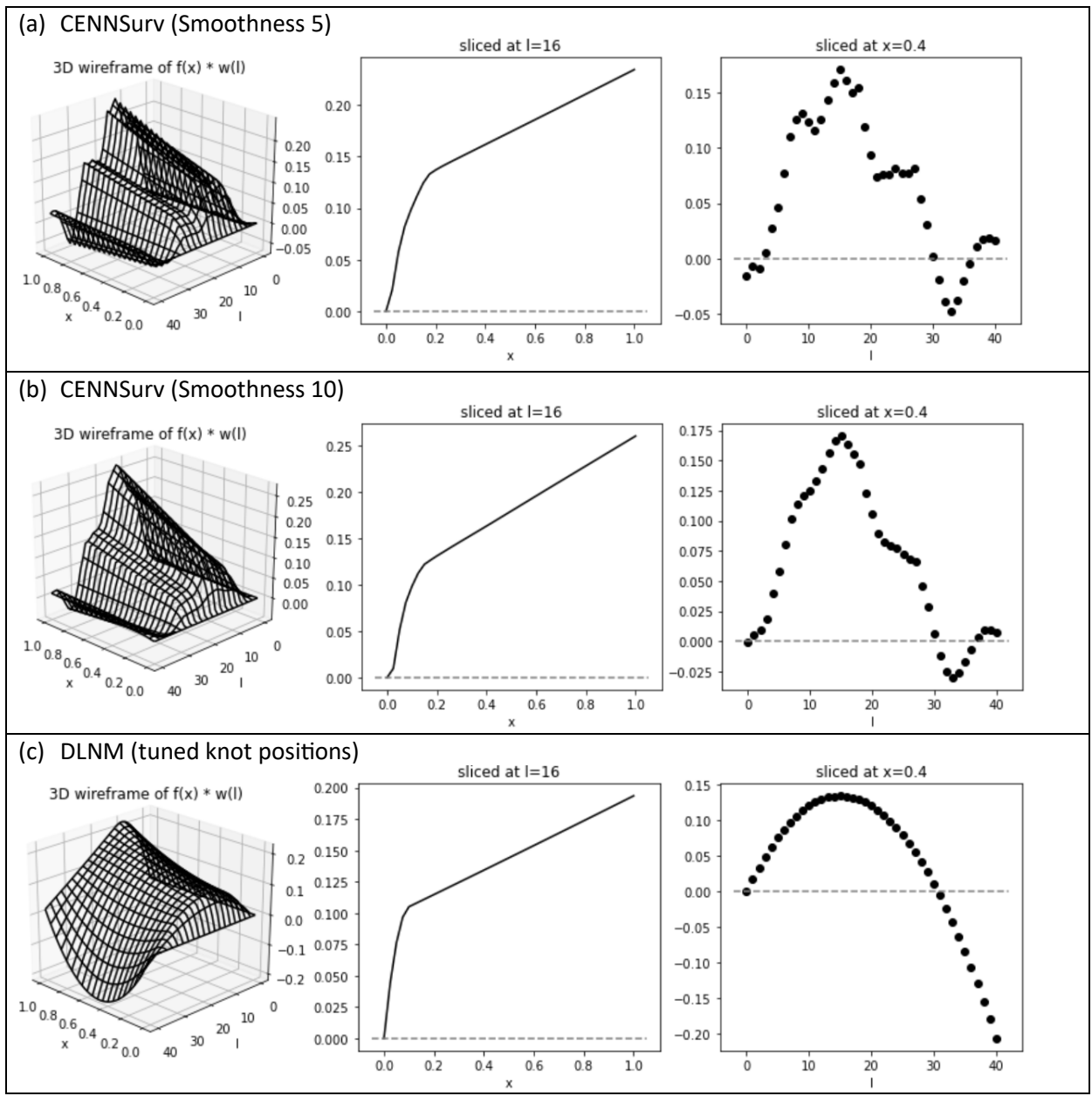

**Supplementary Figure S5**. 3-D Exposure-Lag Contribution Surface and Log Hazard Slice Plots. This figure visualizes the estimated log hazard across different lag periods for three models: CENNSurv with smoothness 5, CENNSurv with smoothness 10, and DLNM with tuned knot positions. The 3-D exposure-lag contribution surfaces for all three configurations display similar cumulative exposure-risk patterns.